\pdfoutput=1

\documentclass[11pt]{article}

\usepackage[final]{acl}

\usepackage{times}
\usepackage{latexsym}

\usepackage[T1]{fontenc}

\usepackage[utf8]{inputenc}

\usepackage{microtype}

\usepackage{inconsolata}

\usepackage{graphicx}

\usepackage[normalem]{ulem}
\useunder{\uline}{\ul}{}
\title{Analyzing Pokémon and Mario Streamers' Twitch Chat with LLM-based User Embeddings}

\author{Mika Hämäläinen\textsuperscript{1}, Jack Rueter\textsuperscript{2} and Khalid Alnajjar\textsuperscript{3} \\
  \textsuperscript{1}Metropolia University of Applied Sciences \\
  \textsuperscript{2}University of Helsinki \\
  \textsuperscript{3}F-Secure Oyj\\
  \texttt{first.last@metropolia.fi/helsinki.fi/fsecure.com} \\}

\begin{document}
\maketitle
\begin{abstract}
We present a novel digital humanities method for representing our Twitch chatters as user embeddings created by a large language model (LLM). We cluster these embeddings automatically using affinity propagation and further narrow this clustering down through manual analysis. We analyze the chat of one stream by each Twitch streamer: SmallAnt, DougDoug and PointCrow. Our findings suggest that each streamer has their own type of chatters, however two categories emerge for all of the streamers: \textit{supportive viewers} and \textit{emoji and reaction senders}. \textit{Repetitive message spammers} is a shared chatter category for two of the streamers.
\end{abstract}

\section{Introduction}

Streamers playing video games online have gained popularity over the past years (see \citealt{alvarez2024twitch}). Platforms like Twitch\footnote{https://twitch.tv} and YouTube\footnote{https://youtube.com} have enabled gamers to build massive audiences, with top streamers attracting millions of followers who tune in regularly to watch live gameplay and to engage in real-time chat (see \citealp{fernandez2024twitch}).

This rise of live-streaming platforms has introduced a novel arena for examining the dynamics of online interaction and community formation. The communicative practices observed in these environments are rich with potential for analysis, offering insights into how digital communities emerge, how chatter identities are performed and negotiated in virtual spaces, and how technology shapes discourse and social interaction (see \citealt{speed2023beyond}).

Real-time chatters provide interesting data for research purposes. In this study, we aim to better understand what categories of chatters are there. We do this by building user embeddings for each chatter using a large language model (LLM) and later on cluster these embeddings into chatter categories. By harnessing the capabilities of large language models (LLMs), we aim to construct a nuanced representation of individual chatters, encapsulating the multifaceted dimensions of their discourse.

We study two Pokémon streams, one by SmallAnt and one by PointCrow, and one Mario stream by DougDoug. We aim to see whether there are any mutually shared chatter categories, and if any streamers has their own unique chatter categories. Through this, we aspire to contribute to a deeper understanding of the sociocultural fabric of online gaming communities by shedding light on the complex interplay between content creators and their audiences in a rapidly evolving digital landscape.

Furthermore, we present a novel embeddings and clustering driven data-analysis method that is applicable as is in almost any digital humanities dataset. The code has been made available on Zenodo\footnote{https://zenodo.org/records/13886601}.

\begin{figure*}[ht]
\centering
\includegraphics[width=1\textwidth]{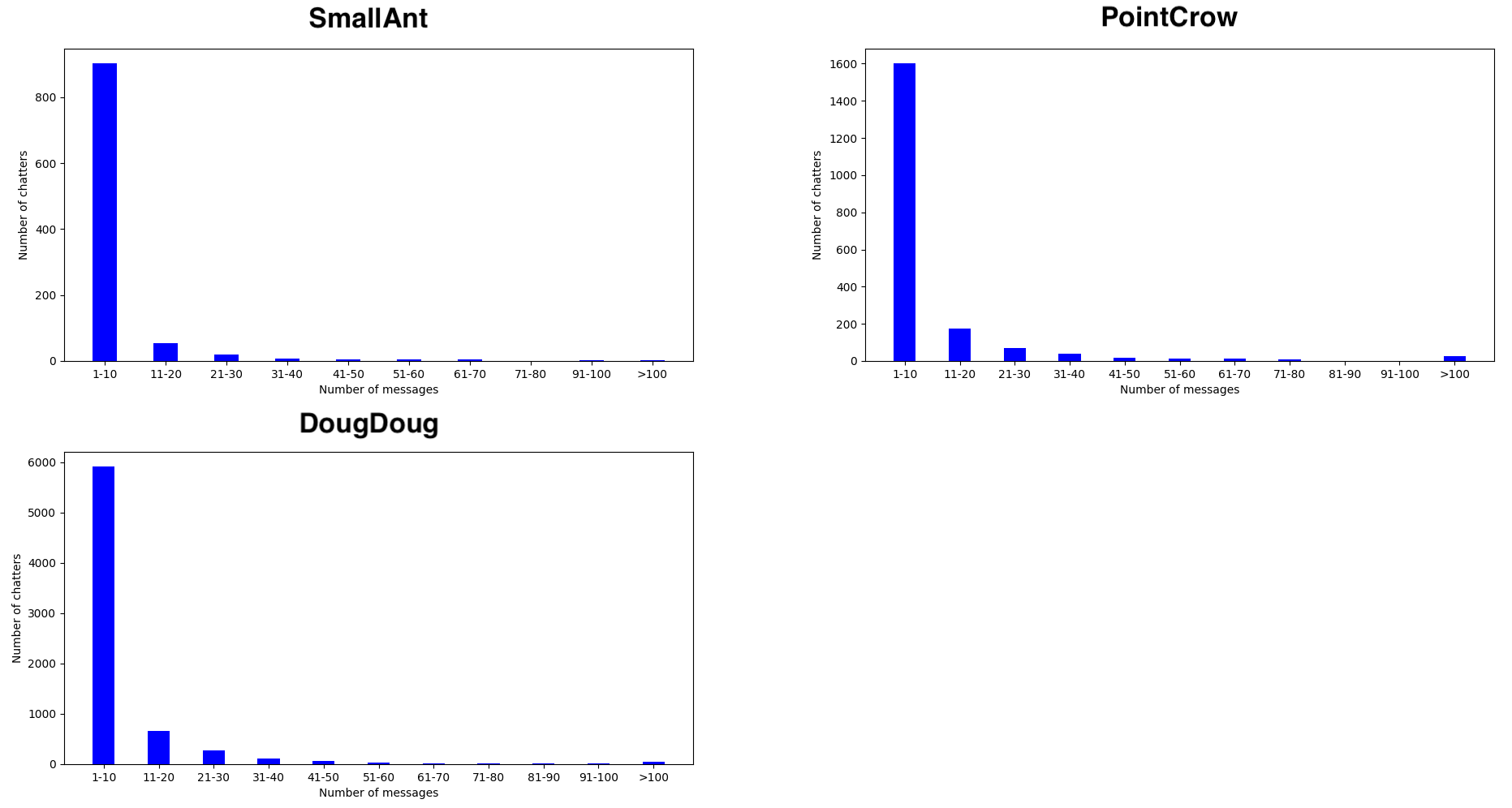}
\caption{Chatter engagement for each stream}
\label{fig:chatternumbers}
\end{figure*}

\section{Related work}

Twitch streams have been an object of study before our research as well. There is also a great body of literature on computational sociolinguistics \cite{saily2021plenipotentiary,tiihonen2023measuring,landert2023tv}. However, in this section we focus on some of the prior literature relating to studying Twitch streamers.

In the context of online live streaming, prior research \cite{recktenwald2017toward} has explored the intricate dynamics between broadcasters and audiences, focusing on the interaction that occurs through both spoken language and written chat during live broadcasts of video games. The research identifies a novel communicative behavior termed 'pivoting,' where both broadcasters and audiences produce context-dependent utterances in response to game events, demonstrating the highly interactive and situated nature of discourse in live streaming environments.

A mixed-methods study \cite{diwanji2020don} explored the information behavior and perceptions of co-presence among Twitch users, leveraging human information theory and social identity theory as the theoretical framework. Through quantitative analysis using tools like LIWC and SPSS, alongside qualitative thematic analysis with Nvivo, the study identified that information reaction and production were prevalent behaviors across multiple streams, while also highlighting the strong sense of co-presence experienced by participants.

The study by \citet{cabeza2022analysis} investigates the complex interplay between video game usage, live streaming and the potential adverse effects associated with excessive engagement in these activities. Employing a multilayer perceptron model on a substantial sample of 970 video game users, the research identifies key factors influencing gaming behavior. Specifically, the study highlights that motivations related to a sense of belonging to gaming platforms and the positive social uses, such as making friends and pursuing gaming as a profession, significantly contribute to the risk of pathological gaming. 

Building on the foundation of affordance theory, a recent study \cite{sjoblom2019ingredients} examined the practices of the most popular streamers, revealing how these individuals leverage various communication modalities and social commerce elements to create compelling content. Their work helps in uncovering the emerging trends and common strategies employed by streamers, offering insights into the evolving landscape of online video streaming as a business model driven by individual content creators.

\section{Twitch chat data}

We study three streams: \textit{HARDEST NUZLOCKE (interrupted by a board game w/ magicthenoah, failboat, captainkidd)}\footnote{https://www.twitch.tv/videos/2220045551} by SmallAnt, \textit{POKEMON FUSION JOHTO LEAGUE 150,000+ NEW FUSIONS | !tts on}\footnote{https://www.twitch.tv/videos/2211596823} by PointCrow and \textit{Can I beat 10 different Mario 64 speedruns simultaneously? !suck}\footnote{https://www.twitch.tv/videos/2217569664} by DougDoug.

Obtaining chat data on Twitch is rather difficult programmatically. We build a small Python script that launches Chrome through Selenium\footnote{https://www.selenium.dev/}. This script monitors the chatbox and saves all chat messages to a file. This requires us to actually watch through the streams to collect the data. The additional benefit is that we do not need to write a crawler that might violate the terms of service of Twitch - this way we simply log out all messages that appear during the streams naturally.

\begin{table}[!ht]
\centering
\begin{tabular}{|l|c|c|c|}
\hline
                   & \multicolumn{1}{l|}{SmallAnt} & \multicolumn{1}{l|}{PointCrow} & \multicolumn{1}{l|}{DougDoug} \\ \hline
Messages & 5088                          & 20488                          & 59207                         \\ \hline
Chatters & 1001                          & 1965                           & 7182                          \\ \hline
Length    & 4h 31min                      & 7h 22min                       & 5h 26min                      \\ \hline
\end{tabular}
\caption{Size of each stream dataset}
\label{tab:datasize}
\end{table}

Table \ref{tab:datasize} shows the key numbers of the dataset size for each streamer. SmallAnt had the lowest viewer engagement out of all the three streamers. DougDoug had the highest number of chatters in his stream.

SmallAnt was playing a Pokémon nuzlocke in his stream where he needed to do some calculations on Pokémon stats. The stream featured a section where he was playing a virtual board game with other streamers. PointCrow's stream was about continuing his playthrough of Pokémon Infinite Fusion game. During the stream, he took a break to eat dinner and watch online videos. DougDoug streamed Super Mario 64. The stream included a section of live coding and a message from his sponsor.

Figure \ref{fig:chatternumbers} shows how active chatters were. Chatters are grouped in categories based on how many messages they sent. As we can see, most of the chatters on all streams sent only 1-10 messages during the entire stream.

\section{User embeddings and clustering}

In order to create a user embedding for each chatter, we concatenate the chat messages for each user and separate each message with a line break. To ensure that we get enough text to represent each chatter well enough, we exclude all chatters who have sent less than 20 messages. If a chatter has not sent enough messages, not enough can be known about them to support the further analysis.

We use the LLM called PaLM-2 \cite{anil2023palm} over Google's VertexAI\footnote{https://cloud.google.com/vertex-ai?hl=en} to embed each chatter's concatenated messages into a user embedding. These user embeddings serve as mathematical representation of the semantics of what each user was chatting about. In particular, we use \textit{text-embedding-004} model for \textit{SEMANTIC\_SIMILARITY} task. We picked this particular task for the embedder because we intend to compare the semantic similarity of each embedding.

We cluster the user embeddings separately for each stream. For clustering, we use affinity propagation \cite{frey2007clustering}. It takes in an affinity matrix, which shows how close each embedding is to other embeddings, and it will automatically find an optimal number of clusters based on the affinities provided to the algorithm. We use cosine similarity to populate the affinity matrix. We use the methods provided in Scikit-learn \cite{scikit-learn} for affinity propagation and cosine similarity. Similar clustering approaches have previously been used with word embeddings \cite{hamalainen2019let,stekel2022word}.

Some chatters end up clustered into their own clusters. We remove all clusters that have only one chatter, because we are more interested in the overall tendencies of chatter categories, not in individual deviant chatters.

\section{Results}

The clustering algorithm created 6 clusters for SmallAnt, 12 clusters for PointCrow and 31 clusters for DougDoug. On a closer inspection, we found that some of the clusters included mutually similar messages, so we proceeded to merge some clusters manually. This resulted in 5 cluster for SmallAnt, 4 for PointCrow and 6 for DougDoug.

\begin{table}[!ht]
\centering
\footnotesize
\begin{tabular}{|l|l|l|}
\hline
Cluster name                                                             & Size & Characteristics                                                                                                                                            \\ \hline
\begin{tabular}[c]{@{}l@{}}Supportive\\ viewers\end{tabular}                 & 8    & \begin{tabular}[c]{@{}l@{}}The chatters are engaged with the stream \\ in a laid-back fashion. The messages are \\ generally positive.\end{tabular}        \\ \hline
\begin{tabular}[c]{@{}l@{}}PartyKirby\\ spammers\end{tabular}            & 7    & \begin{tabular}[c]{@{}l@{}}The chatters were mainly spamming\\ PartyKirby emote\end{tabular}                                                               \\ \hline
\begin{tabular}[c]{@{}l@{}}Emojis and\\ reactions\end{tabular}           & 5    & \begin{tabular}[c]{@{}l@{}}The chatters mainly send emojis and\\ short reactions such as “oh no” or “damn”.\end{tabular}                                   \\ \hline
\begin{tabular}[c]{@{}l@{}}Strategic\\ helpers\end{tabular}              & 11   & \begin{tabular}[c]{@{}l@{}}The chatters sent helpful messages to help \\ the streamer plan their Pokémon party better.\end{tabular}                        \\ \hline
\begin{tabular}[c]{@{}l@{}}Anime and\\ gaming\\ enthusiasts\end{tabular} & 15   & \begin{tabular}[c]{@{}l@{}}These chatters were talking about anime shows\\ and other games as well and not just what \\ SmallAnt was playing.\end{tabular} \\ \hline
\end{tabular}
\caption{SmallAnt chatter clusters}
\label{tab:smant}
\end{table}

\begin{table}[!ht]

\centering
\footnotesize
\hspace*{-1cm}
\begin{tabular}{|l|l|l|}
\hline
Cluster name                                                                     & Size & Characteristics                                                                                                                                                                                 \\ \hline
\begin{tabular}[c]{@{}l@{}}Supportive\\ viewers\end{tabular}                     & 70   & \begin{tabular}[c]{@{}l@{}}The chatters send supportive messages\\ to the streamer. The messages have \\ emotive content and they may also \\ be positive reactions to the stream.\end{tabular} \\ \hline
\begin{tabular}[c]{@{}l@{}}Pokémon Infinite\\ Fusion \\ enthusiasts\end{tabular} & 111  & \begin{tabular}[c]{@{}l@{}}The chatters talk about the game\\ being played and react to different\\ Pokémon fusions with anticipation.\end{tabular}                                             \\ \hline
\begin{tabular}[c]{@{}l@{}}Emojis and \\ reactions\end{tabular}                  & 14   & \begin{tabular}[c]{@{}l@{}}The chatters engage with the stream\\ either by sending emojis or by sending\\ different kinds of short reactions\\ such as “let’s go” or “LMAO”\end{tabular}        \\ \hline
Newcomers                                                                        & 6    & \begin{tabular}[c]{@{}l@{}}These chatters are newcomers\\ to the stream and are watching \\ PointCrow for the first time. Their  \\ messages are generally positive\end{tabular}                 \\ \hline
\end{tabular}
\caption{PointCrow chatter clusters}
\label{tab:pointcrow}
\end{table}

Table \ref{tab:smant} shows the chatter clusters for SmallAnt along with a short description that characterizes the chatters in this category in general terms. The largest category is \textit{Anime and gaming enthusiasts}; this is the only cluster that was merged with a similar cluster that specialized in conversation about a game called \textit{Guilty Gear}.

The results of our analysis on PointCrow's stream can be seen in Table \ref{tab:pointcrow}. All of the clusters required merging except for \textit{Newcomers} cluster. The largest clusters are people who are excited about the game being played, \textit{Pokémon Infinite Fusion}, and \textit{Supportive viewers}.

\begin{table}[]
\centering
\footnotesize
\hspace*{-1cm}
\begin{tabular}{|l|l|l|}
\hline
Cluster name                                                    & Size & Characteristics                                                                                                                                                                                            \\ \hline
\begin{tabular}[c]{@{}l@{}}Emojis and\\ reactions\end{tabular}  & 114  & \begin{tabular}[c]{@{}l@{}}The chatters use emojis and send\\ short reactions that also frequently\\ express confusion such as “huh?” or “D:”\end{tabular}                                                 \\ \hline
\begin{tabular}[c]{@{}l@{}}Meta-level\\ discussion\end{tabular} & 76   & \begin{tabular}[c]{@{}l@{}}The chatters talk about meta-level \\ things relating to DougDoug and his\\ channel such as editors, subscriptions\\ and sponsors.\end{tabular}                                 \\ \hline
\begin{tabular}[c]{@{}l@{}}Critical\\ viewers\end{tabular}      & 188  & \begin{tabular}[c]{@{}l@{}}The chatters criticise DougDoug’s Mario \\ skills. Some of them send !suck command\\ and tell him that the game is rigged.\end{tabular}                                         \\ \hline
\begin{tabular}[c]{@{}l@{}}Supportive\\ viewers\end{tabular}    & 185  & \begin{tabular}[c]{@{}l@{}}The chatters express their support and how\\ much they like DougDoug. Some of the \\ messages may appear negative in tone, \\ but have a clear positive intention.\end{tabular} \\ \hline
Parrots                                                         & 13   & \begin{tabular}[c]{@{}l@{}}These chatters repeat messages with a \\ similar content over and over again. \\ Such as comments about poggies and \\ cheese stream.\end{tabular}                              \\ \hline
\begin{tabular}[c]{@{}l@{}}Random\\ reactions\end{tabular}      & 76   & \begin{tabular}[c]{@{}l@{}}These chatters send plenty of emotive \\ reactions that may be on a variety of \\ different topics.\end{tabular}                                                                \\ \hline
\end{tabular}
\caption{DougDoug chatter clusters}
\label{tab:dougdoug}
\end{table}

Table \ref{tab:dougdoug} shows the clusters of DougDoug chatters. \textit{Parrots} is the only cluster that did not require manual merging. The largest number of merged clusters are in \textit{Critical Viewers} (11 clusters) and in \textit{Supportive viewers} (9 clusters). Interestingly DougDoug has many chatters that send negative comments. In fact, there is an active Reddit r/wehatedougdoug\footnote{https://www.reddit.com/r/wehatedougdoug/}, which indicates that hatered is a valid way of showing DougDoug fandom. Perhaps this polarity between the supporters and supposed haters is the reason why DougDoug has the highest number of active chatters out of all the streamers.

If we look at all the chatter cluster for all the streamers, we can see that a number of chatter groups emerge for all the streamers: \textit{Supportive viewers} and \textit{Emojis and reactions}. In addition, SmallAnt and DougDoug have a cluster of people who spam repetitive messages (such as PartyKirby).  

\section{Conclusions}

In conclusion, our study introduces a novel approach within digital humanities by utilizing a large language model (LLM) to create user embeddings for representing Twitch chatters. By employing affinity propagation for automatic clustering and refining the results through manual analysis, we were able to effectively categorize chat participants from streams by SmallAnt, DougDoug, and PointCrow.

The method is generic enough to be used with any kinds of documents. In our study, we found that a bit of manual merging of the clusters was needed. This can be automatized in the future by running multiple iterations of the clustering algorithm by clustering clusters. This could be achieved easily by calculating a centroid embedding for each cluster and recalculating the affinity matrix based on the cluster centroids.

Our analysis revealed that, while each streamer attracts a distinct type of chatters, there are common categories across the streams. Notably, all three streamers share categories of supportive viewers and emoji and reaction senders. Additionally, a category of repetitive message spammers was found to be common among two of the streamers. These insights highlight the potential of LLM-based embeddings for understanding and categorizing social interactions in digital environments.

In the future, it might be interesting to gather more chat messages on multiple streams of the same streamer to assess the stability of the chatter categories across different streams. Chat messages also include timestamps, which we did not take into consideration in this study. It might also be interesting to include the content of the actual stream in the study, as our current study was only limited to the chat messages. 

\section{Limitations}

The sample size for this paper is relatively small as it only covers one stream from three streamers. This means that the findings will not necessarily hold for every Twitch streamer or other streams by the three streamers that were studied.



\bibliography{acl_latex}

\end{document}